# A Novel Approach Towards Clustering Based Image Segmentation

**Dibya Jyoti Bora, Anil Kumar Gupta**

*Abstract— In computer vision, image segmentation is always selected as a major research topic by researchers. Due to its vital rule in image processing, there always arises the need of a better image segmentation method. Clustering is an unsupervised study with its application in almost every field of science and engineering. Many researchers used clustering in image segmentation process. But still there requires improvement of such approaches. In this paper, a novel approach for clustering based image segmentation is proposed. Here, we give importance on color space and choose l\*a\*b\* for this task. The famous hard clustering algorithm K-means is used, but as its performance is dependent on choosing a proper distance measure, so, we go for "cosine" distance measure. Then the segmented image is filtered with sobel filter. The filtered image is analyzed with marker watershed algorithm to have the final segmented result of our original image. The MSE and PSNR values are evaluated to observe the performance.*

## I. INTRODUCTION

Image segmentation is the process of partitioning an image into uniform and non overlapping regions in order to find out meaningful information from the segmented images. These regions can be considered homogeneous according to a given criterion, such as color, motion, texture, etc. Image segmentation is always the researcher's first choice due to its leading rule in image processing research. Clustering is an unsupervised study where data items are partitioned into different groups known as clusters by keeping in mind two properties: (1) High Cohesion and (2) Low Coupling [1]. According to the first property, data items belong to one particular cluster must show high similarities. And, the second property says that data items of one cluster should be different from the data items of the other clusters. Clustering is divided into two main types: Hard Clustering (Exclusive Clustering) and Soft Clustering (Fuzzy Clustering or Overlapping Clustering)[2]. In hard clustering, data items are clustered in an exclusive way, so that if a particular data item belongs to a definite cluster then it could not be included in another cluster. While, in case of soft clustering, data items may exhibit membership values to more than one cluster. Clustering is suitable for image segmentation task [3]. In our approach, we have selected K-Means algorithm which is a famous hard clustering algorithm due to its low computational complexity [2]. Choosing a proper distance measure always affects the clustering result [4]. Generally, researchers use to choose "Euclidean" as distance measure.

**Mr. Dibya Jyoti Bora**, Currently Teaching PG Students in the Department of Computer Science & Applications, Barkatullah University, Bhopal, India.

**Dr. Anil Kumar Gupta**, HOD of the Department of Computer Science & Applications, Barkatullah University, Bhopal, India.

But, in our approach we have chosen "Cosine" as our distance measure. Then, when it comes to color image segmentation, choosing a proper color space is a mandatory condition to optimize the result [5]. The l\*a\*b\* color space is selected for our experiment due to its better performance in color image segmentation [6]. We first converted the rgb image into l\*a\*b\* color space. Then, the resultant image is clustered with K-Means algorithm with cosine distance measure. Now, gradient magnitude of the segmented image is calculated with sobel filter. After that, the resultant image is segmented with marker controlled watershed method. At last, we have calculated the MSE and PSNR values of the original image with respect to the final segmented image to observe the accuracy of the proposed method. The paper is designed as follows: first of all, a short review on previous work done in the field is given. Then, the topics concerned with the approach are illustrated clearly, followed by the flowchart of the proposed approach. After that, the experiment's results are shown with concerning figures. At last, a conclusion is drawn with giving an idea about future research work that may be possible in the field.

## II. REVIEW ON PREVIOUS WORK DONE IN THE FIELD

In recent image processing research, color image processing is getting high preference due to the reason that human eyes are more adjustable to brightness , so, we can identify thousands of colors at any point of a complex image, while only a dozens of gray scale are possible to be identified at the same time[7][8]. In [7], the authors proposed a color image segmentation method based on watershed algorithm with a combination of seed region growing algorithm. In [9], the authors proposed an image segmentation technique where adaptive wavelet thresholding is used for de-noising followed by Marker controlled Watershed Segmentation. In [10], an improved watershed algorithm is proposed where median filters are used for de-noising in order to enhance the performance of watershed algorithm. In [11], authors try to compare the performance of integrated k-means algorithm and watershed algorithm with integrated fuzzy c means algorithm and watershed algorithm. In [12], authors transformed the image from rgb color space to l\*a\*b\* colorspace. After separating three channels of l\*a\*b\*, a single channel is selected based on the color under consideration. Then, genetic algorithm is applied on that single channel image. The method is found to be very effective in segmenting complex background images. In [13] , the image is converted from rgb to l\*a\*b\* colorsace, and then  k means algorithm is applied to the resulting image. Then pixels are labeled on the segmented image. Finally images are created which segment the original image by color.

### III. L*A*B* COLOR SPACE

This color space is defined by CAE and specified by the International Commission on Illumination [14][15]. Here, one channel is for Luminance (Lightness) and other two color channels are a and b known as chromaticity layers. The* layer indicates where the color falls along the red-green axis, and the another chromaticity layer b* indicates where the color falls along the blue-yellow axis. a* negative values indicate green while positive values indicate magenta; and b* negative values indicate blue and positive values indicate yellow. One of very important characteristic of this color space is that this is device independent [16], means to say that this provides us the opportunity to communicate different colors across different devices. Following figure clearly illustrates the coordinate system of l*a*b* color space [17]:

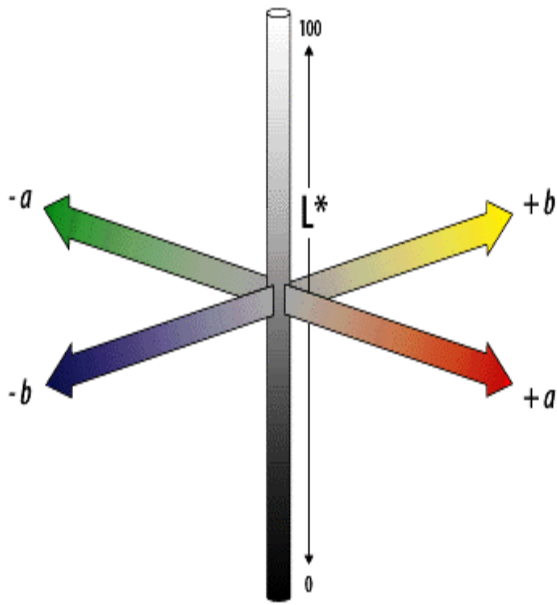

**Fig. (1) L*A*B* Color Space**

In the above figure, the central vertical axis represents lightness (L*). L* can take values from 0(black) to 100(white). Here, the fact that "a color cannot be both red or green, or both blue and yellow, because these colors oppose each other" is obeyed by the co ordinate axes. For each axis, values run from positive to negative. Positive 'a' values indicate amounts of red, while negative values indicate amounts of green. And, positive 'b' values indicate amount of yellow while, negative 'b' values indicate blue. And zero represents neutral gray for both the axes. So, in this color space, values are needed only for two color axes and for the lightness or grayscale axis (L*).

### IV. K-MEANS ALGORITHM

Among the hard clustering algorithms, K-Means algorithm [18][19] is always researcher's first choice because of its simplicity and high performance ability. Here, K is the numbers of clusters to be specified. The formal steps involved in this algorithm are:

Step 1: Choose K initial centroids

Step2: Determine the distance of each data item to the centroid.

Step 3: Form k clusters by assigning the data items to the closest centroid.

Step 4: Re compute the centroid of each cluster.

Step 5: Repeat step 3 and step 4 until the centroids do not change their positions.

The aim of the K-Means algorithm is to minimize the squared error function [4]:

$$J = \sum_{j=1}^{k} \sum_{i=1}^{n} \left\| x_i^{(j)} - c_j \right\|^2$$

Here, $\left\| x_i^{(j)} - c_j \right\|^2$ is a chosen distance measure between a data point $x_i^{(j)}$ and the cluster centre $c_j$.

### V. COSINE DISTANCE

The cosine distance between two points is one minus the cosine of the included angle between points (treated as vectors). This distance is based on the measurement of orientation and not magnitude [20]. Given an $m$-by-$n$ data matrix X, which is treated as $m$ (1-by-$n$) row vectors $x_1$, $x_2$, ..., $x_m$, the cosine distances between the vector $x_s$ and $x_t$ are defined as follows[4] :

$$d_{st} = 1 - \frac{x_s x_t'}{\sqrt{(x_s x_s')(x_t x_t')}}.$$

We have chosen this distance measure because of its devotion towards orientation of data points which is much more concerned to our work while dealing with pixels of an image.

### VI. SOBEL FILTER

The Sobel filter (also, known as Sobel operator) ,named after Irwin Sobel[21], is generally used in edge detection algorithms to create an image by emphasizing edges and transitions of the image. This is a discrete differentiation operator which computes an approximation of the gradient of the image intensity function. This is based on convolving the image with a small, separable, and integer valued filter in horizontal and vertical direction and is therefore inexpensive in terms of computations [22]. Actually, this is an orthogonal gradient operator [23], where gradient corresponds to first derivative and gradient operator is a derivative operator. For an image, here involves two kernels: Gx and Gy ; where Gx is estimating the gradient in x-direction while Gy estimating the gradient in y-direction. Then the absolute gradient magnitude will be given by:

$$|G| = \sqrt{(Gx^2 + Gy^2)}$$

Although, this value is often approximated with [22][23] :

$$|G| = |Gx| + |Gy|$$

In our work, we choose sobel operator because of its capacity of smoothing effect on the random noises of an image[24]. Since it is differentially separated by two rows and columns, so the edge elements on both sides become enhanced which offer a very bright and thick look of the edges.

### VII. MARKER-CONTROLLED WATERSHED SEGMENTATION

This is a region-based image segmentation process, first proposed by Digabel and Lantuejoul[25]. The concept behind this algorithm is originally drawn from the geographical concept of watershed[26][27]. A watershed is the ridge that divides areas drained by different river systems. If we view an image as a geological landscape, the watershed lines determine the boundaries that separate image regions. The numerical value (i.e., the gray tone) of each pixel of an image in topographic representation stands for the evolution at this point. Then, the watershed transform computes the catchments basins and ridge lines, with catchment basins corresponding to image regions and ridge lines relating to region boundaries [28]. Marker-controlled watershed segmentation is a potent and flexible method for segmentation of objects with closed contours, where the extremities are expressed as ridges. The method consists of the following steps [29]:

1. First of all, compute a segmentation function (this is an image whose dark regions are the objects we are trying to segment).

2. Compute foreground markers which are the connected blobs of pixels within each of the objects.

3. Compute background markers which are pixels that are not part of any object.

4. Then modify the segmentation function so that it only has minima at the foreground and background marker locations.

5. At last, compute the watershed transform of the modified segmentation function.

## VIII. MSE AND PSNR

The MSE (Mean Squared Error)is the cumulative squared error between the compressed and the original image, whereas PSNR(Peak Signal to Noise Ratio) is a measure of the peak error[31][32]. The formula for MSE[30] is:

$$MSE = \frac{1}{MN} \sum_{y=1}^{M} \sum_{x=1}^{N} \left[ I_{(x,y)} - I'_{(x,y)} \right]^2$$

where, I(x,y) is the original image, I'(x,y) is its noisy approximated version (which is actually the decompressed image) and M,N are the dimensions of the images. A lower value for MSE implies lesser error.

The formula for PSNR[30] is

$$PSNR = 10 \cdot \log_{10} \left( \frac{MAX_I^2}{MSE} \right)$$

Where, $MAX_I$ is the maximum possible pixel value of the image. A higher value of PSNR is always preferred as it implies the ratio of Signal to Noise will be higher. The 'signal' here is the original image, and the 'noise' is the error in reconstruction.

## IX. FLOWCHART OF THE PROPOSED APPROACH

The flow chart of our proposed can be presented as follows:

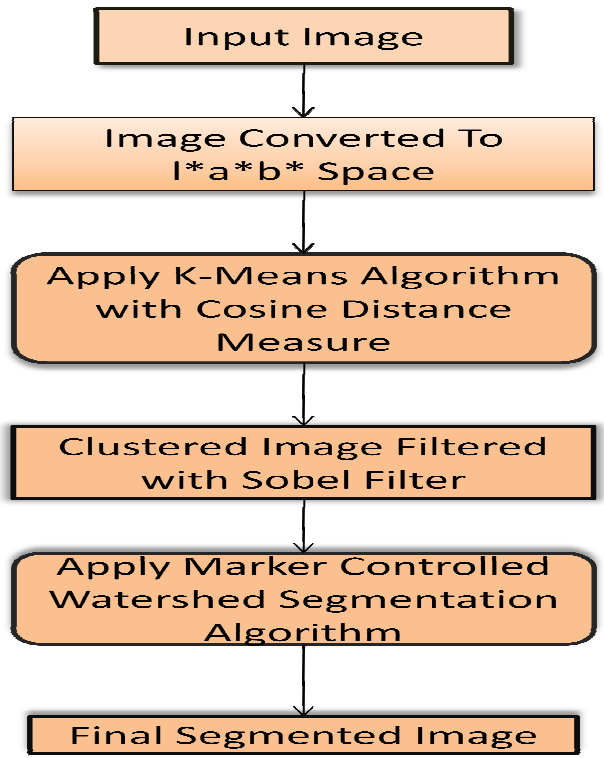

**Fig. (2) Proposed Approach**

## X. EXPERIMENTS

For our experiment, we have chosen Matlab. Then, we applied our proposed approach on the 'onion' image that is available as Matlab demo image [33]. We initialized K (number of clusters) value as 3 for the K-Means algorithm. Following are the series of images that result from our experiment:

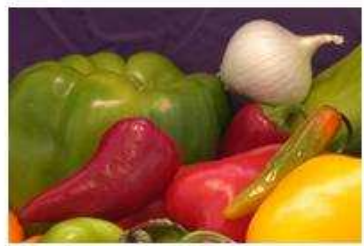

**Fig. (3) Original Image**

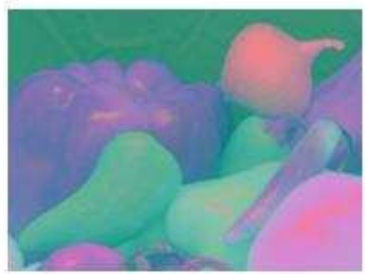

**Fig (4): L\*A\*B\* Converted Image**

A Novel Approach Towards Clustering Based Image Segmentation

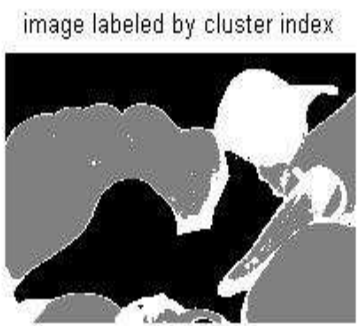

**Fig. (5) Image Obtained after K-Means with Cosine Distance Measure**

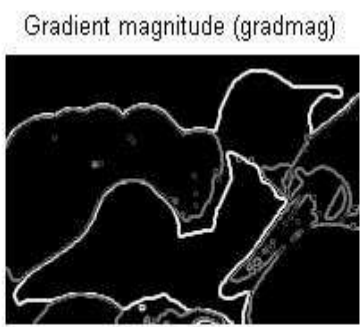

**Fig. (6) Image Obtained after Filtered by Sobel Filter**

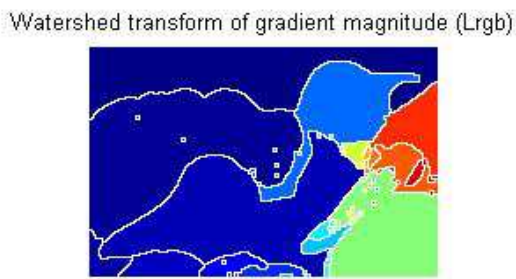

**Fig. (7) Final Segmented Image after Watershed Transform**

At last, we have calculated the MSE and PSNR values from the original and final segmented image (shown in table (1)).

| MSE Value | PSNR Value |
|---|---|
| MSE(:,:,1) : 557110.37 | PSNR(:,:,1): |
| MSE(:,:,2) : 171336.89 | -9.2946131 dB |
| MSE(:,:,3) : 3915.66 | PSNR(:,:,2) : |
| | -4.1737095 dB |
| | PSNR(:,:,3): |
| | 12.2367546 dB |

**Table (1): Table Showing the Values of MSE and PSNR Values Calculated from Original and Final Segmented Image**

So, from the final segmented image, it is clearly visible that our proposed approach succeeds to produce an enhanced segmentation of the original color image. Also, from table(1), if we observe the MSE and PSNR values, then the result can be said quite satisfactory in relation to the fact that a lower MSE value and a higher PSNR value leads to a better segmentation.

## XI. CONCLUSION

In this paper, we propose a novel approach for clustering based image segmentation. We consider color image segmentation instead of gray image segmentation because of the former's capability to enhance the image analysis process thereby improving the segmentation result. The proposed approach produces a very effective result of color image segmentation. But, here, we need to specify the number of clusters beforehand. So, an incorrect assumption may somewhat make the final segmented image blurry. Hence, in our future research, we will deal with this fact so that this proposed novel approach can further be improved with a good determined number of clusters.

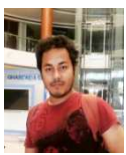
**Mr. Dibya Jyoti Bora**, Ph.D. in Computer Science pursuing, M.Sc. in Information Technology (University 1st Rank holder Barkatullah University, Bhopal), GATE CS/IT Qualified, GSET qualified in Computer Science, Distinction holder in graduation (honors in Mathematics, Dibrugarh University, Assam). Currently teaching PG students in the Department of Computer Science & Applications, Barkatullah University, Bhopal. Research interests are Cluster Analysis and its applications in Image Processing and Machine Learning.

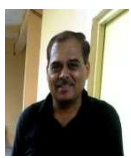
**Dr. Anil Kumar Gupta**, Ph.D.in Computer Science (Barkatullah University, Bhopal), HOD of the Department of Computer Science & Applications, Barkatullah University, Bhopal. Research interest: Data Mining, Artificial Intelligence and Machine Learning.